%% file: main.tex
\def\year{2019}\relax
\documentclass[letterpaper]{article}
\usepackage{aaai19}
\usepackage{times}
\usepackage{helvet}
\usepackage{courier}
\usepackage{url}
\usepackage{graphicx}
\usepackage{hyperref}
\usepackage{amsmath}
\usepackage{amssymb}
\usepackage{booktabs} 
\usepackage{xcolor}
\usepackage{algorithm}
\usepackage[noend]{algpseudocode}
\usepackage{comment}

\newcommand{\citet}[1]{\citeauthor{#1} \shortcite{#1}}

\title{Deep Policies for Width-Based Planning in Pixel Domains$^*$}
\author{
Miquel Junyent,\textsuperscript{\rm 1}
Anders Jonsson,\textsuperscript{\rm 2}
Vicen\c{c} G'{o}mez,\textsuperscript{\rm 2}\\
\textsuperscript{\rm 1}Grupo Oes\'{i}a,
\textsuperscript{\rm 2}Universitat Pompeu Fabra\\
mjunyent@oesia.com, anders.jonsson@upf.edu, vicen.gomez@upf.edu
}
\author{Miquel Junyent \\
Grupo Oes\'{i}a \\
Barcelona, Spain \\
mjunyent@oesia.com
\And Anders Jonsson \\
Universitat Pompeu Fabra \\
Barcelona, Spain \\
anders.jonsson@upf.edu
\And Vicen\c{c} G\'{o}mez \\
Universitat Pompeu Fabra \\
Barcelona, Spain \\
vicen.gomez@upf.edu
}

\begin{document}

\maketitle
\begin{abstract}

Width-based planning has demonstrated great success in recent years due to its ability to scale independently of the size of the state space. For example, Bandres et al. (2018) introduced a rollout version of the Iterated Width algorithm whose performance compares well with humans and learning methods in the pixel setting of the Atari games suite. In this setting, planning is done on-line using the ``screen" states and selecting actions by looking ahead into the future. However, this algorithm is purely exploratory and does not leverage past reward information. Furthermore, it requires the state to be factored into features that need to be pre-defined for the particular task, e.g., the B-PROST pixel features. In this work, we extend width-based planning by incorporating an explicit policy in the action selection mechanism. Our method, called $\pi$-IW, interleaves width-based planning and policy learning using the state-actions visited by the planner. The policy estimate takes the form of a neural network and is in turn used to guide the planning step, thus reinforcing promising paths. Surprisingly, we observe that the representation learned by the neural network can be used as a feature space for the width-based planner without degrading its performance, thus removing the requirement of pre-defined features for the planner.
We compare $\pi$-IW with previous width-based methods and with AlphaZero, a method that also interleaves planning and learning, in simple environments, and show that $\pi$-IW has superior performance.
We also show that $\pi$-IW algorithm outperforms previous width-based methods in the pixel setting of Atari games suite.

\end{abstract}

\section{Introduction}\label{sec:intro}

Width-based search algorithms have recently emerged as a state-of-the-art approach to automated planning~\cite{Lipovetzky2012WidthProblems}. These algorithms assume that states are factored into {\em features} and rely on the concept of {\em state novelty}, which measures how novel a state is with respect to the states already visited during search. More precisely, a state is novel if at least one tuple of feature values appears for the first time during search, otherwise the state is pruned. The {\em width} of an algorithm bounds the size of the tuples used for novelty tests. Crucially, the complexity of width-based algorithms is exponential in the width, but independent of the size of the state space.

The defining trait of width-based algorithms is the ability to perform structured exploration of the state space, quickly reaching distant states that may be important for achieving planning goals. In classical planning, width-based search has been incorporated in a heuristic search framework, with algorithms such as Best-First Width Search~\cite{Lipovetzky2017Best-FirstPlanning} achieving remarkable results at the 2017 International Planning Competition. 

Width-based search has also been adapted to reward-driven problems with unknown dynamics, relying on a simulator for interaction with the environment~\cite{Frances2017PurelySimulators}. This enables width-based search in domains intended for reinforcement learning, such as the Arcade Learning Environment~\cite{Bellemare2012TheAgents}. Several researchers have adapted the Iterated Width (IW) algorithm to this setting, using the bytes of the internal RAM state as features~\cite{Lipovetzky2015ClassicalGames,shleyfman2016blind,jinnai2017learning}. Width-based search is used to select the action that maximizes long-term reward, and search is restarted after each action execution.

Since humans do not have access to the RAM state when playing a game, learning algorithms instead define the state in terms of the pixels of the game screen. Recently, \citet{bandres2018planning} proposed a modified version of IW, Rollout IW, that uses pixel-based features and achieves comparable results to learning methods in almost real-time. The main bottleneck of IW is the cost of resetting the simulator to a previous state, and Rollout-IW gets around this by resetting the state only once before each rollout.

In spite of these achievements, IW remains a purely exploratory algorithm that is highly dependent on the quality of the given features and that expands actions at random. Since IW is sensitive to the order of action expansion, there is an opportunity to perform a more informed action selection.

In this work we leverage recent progress in deep reinforcement learning to train a {\em policy} $\pi$ in the form of a neural network (NN) whose inputs are the pixels of an image, such as the Atari game screen. The policy is trained on the actions output by IW, reinforcing action trajectories that have proven promising in the past. The resulting algorithm, $\pi$-IW, uses the policy $\pi$ to select actions during width-based search, resulting in more informed exploration of the state space.

We also show that the last intermediate layer of the learned policy NN can be used as features for IW. Surprisingly, such dynamic features are often competitive with hand-crafted features, maintaining or sometimes increasing the performance of IW. In some domains, we even show that the dynamic features effectively reduce the width of a problem. Experimental results indicate that our approach to pixel-based planning is highly competitive in the Atari suite.

\section{Background}

In this section, we review the fundamental concepts of width-based planning and Markov decision processes.

\subsection{Iterated Width}

Iterated Width (IW) is a pure exploration algorithm originally developed for goal-directed planning problems with deterministic actions~\cite{Lipovetzky2012WidthProblems}. It requires the state space to be factored into a set of {\em features} 
$\Phi$. All features are usually assumed to have the same domain $D$, e.g.~binary ($D=\{0,1\}$) or integer ($D=\mathbb{Z}$). The original algorithm consists of a sequence of calls IW($i$) for $i=0,1,2,\hdots$ until a termination condition is reached.
IW($i$) performs a standard breadth-first search (BrFS) from a given initial state~$s_0$, but prunes states that are not {\em novel}.
When a new state $s$ is generated, IW($i$) contemplates all $n$-tuples of atoms of $s$ with size $n \leq i$. The state is considered novel if at least one tuple has not appeared in the search before, otherwise it is pruned. This notion of novelty has been extended to produce several variants of the IW algorithm, e.g.~those explained in the next sections.

IW($i$) is thus a $\emph{blind}$ search algorithm that eventually traverses the entire state-space if we make $i$ large enough. The traversal depends on how the states are structured, i.e., which features are used to represent the states, and on the order in which states are expanded.
Each iteration IW($i$) is an $i$-width BrFS 
that is complete for problems whose width is bounded by $i$ and visits at most $\mathcal{O}((|\Phi|\cdot|D|)^i)$ states, where $|\Phi|$ is the number of features and $|D|$ is the size of their domains~\cite{Lipovetzky2012WidthProblems}.
Interestingly, most planning benchmarks turn out to have very small width and, in practice, they can be solved in linear or quadratic time.

\subsection{Markov Decision Processes}

A Markov decision process (MDP) is a tuple $M = \langle S,A,P,r \rangle$, where $S$ is the finite state space, $A$ is the finite action 
space, $P:S \times A \rightarrow \Delta(S)$ is the transition function,
and $r:S \times A \rightarrow \mathbb{R}$ is the reward function. 
Here, $\Delta(S)=\{\mu\in\mathbb{R}^{S}: \sum_s\mu(s)=1, \mu(s)\geq 0 \; (\forall s)\}$ is the probability simplex over $S$. 
Although MDPs are stochastic, planning algorithms typically assume that transitions are {\em deterministic}; this is the case for our algorithm as well.

At time $t$, the learner observes state $s_t \in S$, selects action $a_t \in A$, moves to the next state $s_{t+1} \sim P(\cdot|s_t,a_t)$, and obtains reward $r_{t+1}$ such that $\mathbb{E}[r_{t+1}]=r(s_t,a_t)$.
The aim of the learner is to select actions that maximize the expected cumulative discounted reward $\mathbb{E}\left[\sum_{k=t}^\infty \gamma^{k-t} r_{k+1}\right]$, where $\gamma\in(0,1]$ is a discount factor. 
In the reinforcement learning setting, both the transition function $P$ and the reward function $r$ are unknown to the learner.
Instead, the learner relies on a simulator in order to sample both functions.

The decision strategy is represented by a {\em policy} $\pi:S \rightarrow \Delta(A)$, i.e.~a mapping from states to probability distributions over actions. In state $s_t$, action $a_t$ is selected with probability $\pi(a_t|s_t)$.
In a realistic setting, the state space is too large to explicitly represent the policy for each state. Instead, the policy estimate $\widehat\pi_\theta$ maintains a set of {\em parameters} $\theta$ that are combined with the state features to produce a distribution over actions. It is common to use an NN to represent the policy estimate, and define $\widehat\pi_\theta(a|s) = \frac{\exp(h_a(s,\theta)/\tau)}{\sum_{b \in A}{\exp(h_b(s,\theta)/\tau)}}$ as the softmax combination of the NN outputs $h_a(s,\theta)$, $a\in A$, where $\tau$ is a temperature that controls exploration.

\section{Related Work}{
Our work lies at the intersection between width-based planning and policy learning in RL. We now review some relevant work of these two lines of research.

\subsection{Width-Based Planning for MDPs}\label{sec:rw}
Several extensions of the original IW algorithm have been developed over the last years. One line of research has focused on the popular Atari 2600 benchmark~\cite{Bellemare2012TheAgents}, predominantly 
in the on-line planning setting, where actions are selected after a lookahead, and using the 128 bytes of the RAM memory to represent the state of the game.

First,~\citet{Lipovetzky2015ClassicalGames} extended the original IW algorithm to MDPs by associating a reward $R(s)$ to each state $s$ during search, equivalent to the reward $\sum_{t=0}^{d-1} \gamma^t r_{t+1}$ accumulated on the path from $s_0$ to $s_d=s$, where $d$ is the depth of $s$ in the search tree. Later,~\citet{shleyfman2016blind} introduced prioritized-IW (p-IW) that approximates breadth-first search with duplicate detection and state reopening.~\citet{jinnai2017learning} further extended p-IW by learning to avoid actions that lead to the same successor state. Both extensions considerably outperformed the original IW in the RAM setting of the Atari benchmark.

More recently,~\citet{bandres2018planning} adapted IW for the \emph{pixel setting}. Specifically, their method uses the (binary) B-PROST features~\cite{Liang2016StateLearning}. The authors introduced a Rollout version of IW(1) which \emph{emulates} the breadth-first traversal of IW, by keeping track of the minimum depth at which a feature is found for the first time and extending the notion of novelty accordingly. The pruned states are kept as leaves of the tree, and are considered as candidates for states with highest reward. This contribution has brought width-based planning closer to the RL setting. For example, Rollout-IW can deal with trajectories, as most RL methods do, with the restriction that a simulator needs to be reset to a previous state. 
Furthermore, the algorithm can work directly with pixels, instead of the RAM states, which are not always accessible.
This Rollout version of IW is the basis of our work.

Besides the developments on the Atari benchmark, other works consider the discovery of features of IW and possible methods to inform the purely exploratory search of the original IW algorithm. For example,~\citet{Frances2017PurelySimulators} discovered numeric features by means of a IW(1) search by keeping track of trajectories reaching a state where at least one of the goals in the task is true. Also, in~\citet{Lipovetzky2017Best-FirstPlanning}, an~\emph{informed} variant of IW was proposed where a standard goal-oriented search (exploitation) and width-based search (structural exploration) are combined to yield a search scheme, best-first width search, that results in classical planning algorithms that outperform many state-of-the-art planners.

These latter contributions have been developed in the classical planning setting. Our interest is to exploit the potential of deep learning methods for learning representations and policies directly from screen states by leveraging the structured exploration benefits of width-based planning.

\subsection{Policy Iteration for Multi-Step Lookahead Policies}
A natural way to combine planning and learning is to treat the planner as a ``teacher" that provides~\emph{correct} transitions that are used to learn a policy, as in imitation learning~\cite{ross2011reduction,guo2014deep}. A prominent example of this approach is AlphaGo, which achieved superhuman performance in the game of Go \cite{silver2016mastering} by combining supervised learning from expert moves and self-play. AlphaZero~\cite{Silver2017MasteringKnowledge}, a version of the same algorithm that learned solely from self-play, has outperformed previous variants, also showing stunning results in Chess and Shogui~\cite{Silver2017MasteringChess}. This combination of planning and learning can also be interpreted as a policy iteration algorithm with policy updates that
consider multi-step (lookahead) greedy policies~\cite{efroni2018beyond}, instead of the classical 1-step greedy policy improvement with respect to a value function estimate~\cite{sutton1998reinforcement}.

At every iteration $t$, AlphaZero generates a tree using Monte-Carlo tree search (MCTS)~\cite{browne2012survey}, guided by a policy and a value estimate. It keeps a visit count on the branches of the tree, and uses it to explore less frequent states~\cite{kocsis2006bandit} and to generate a target policy $\pi^{target}_t$. After tree expansion, an action is selected at the root following $\pi^{target}_t$, and the resulting subtree is kept for the next iteration. At the end of the episode, the win/loss result $z_t$ is recorded and all transitions $(s_t, \pi^{target}_t, z_t)$ are added to a dataset. In parallel, the policy and value estimates are trained in a supervised manner with minibatches sampled from the dataset.

AlphaZero is known to perform well in two-player games. However, not much is known about its performance on general sequential decision problems. Here we show limitations of AlphaZero in simple environments. We postulate that its poor performance can be caused by the (unstructured) exploration of MCTS, in the sense that MCTS considers the state representation as a black-box.


}

\section{Policy-Guided Iterated Width ($\pi$-IW) }\label{sec:approach}

In spite of its success, IW does not learn from experience, so its performance does not improve over time.
When expanding a node, IW generates all possible states, one per action.
The recently proposed Rollout IW algorithm~\cite{bandres2018planning} generates whole branches by expanding one state per node, but selects actions randomly. 

We now present our algorithm, Policy-Guided Iterated Width ($\pi$-IW), that enhances Rollout IW by incorporating an action selection policy, resulting in an \emph{informed} IW search. More precisely, we leverage the exploration capacity of IW to train a policy estimate $\widehat\pi_\theta$, which is used in turn to guide subsequent search. We consider tuples of size $1$, i.e., IW(1), which keeps planning tractable. 
Similar to Rollout~IW, $\pi$-IW requires a simulator that provides the successor of a state $s$ and a representation of $s$ in terms of features $\Phi$.
Also, $\pi$-IW operates in an online setting, i.e., at each time-step, a planning step is followed by an action execution step.
Importantly, we reset the novetly table after each action step, which enables us to solve problems of width higher than one.


Below we describe the two basic steps of the $\pi$-IW algorithm, and present a mechanism for extracting a feature space from the policy $\widehat\pi_\theta$. This second use of the policy is beneficial if no feature representation is initially available.

\subsection{Planning Step}




The planning step of $\pi$-IW is very similar to Rollout IW. For clarity, we describe four separate functions in Algorithm~\ref{algo}\footnote{The full code is available at \href{https://github.com/aig-upf/pi-IW}{https://github.com/aig-upf/pi-IW}.}. At every iteration of \textit{Lookahead}, Rollout IW first selects a node $n$ for expansion, and then performs a rollout from $n$. \textit{Select} samples actions to traverse the tree until a state-action pair $(n, a)$ is reached that has not yet been expanded. \textit{Rollout} then samples actions starting from $(n, a)$ until a state is reached that is either terminal or not novel. At that point, the final node is marked as \textit{solved} and the process restarts until all nodes have been solved or a maximum budget of time or nodes is exhausted.



Following \citet{bandres2018planning}, a state is considered novel if one of its atoms is true at a smaller depth than the one registered so far in the novelty table. A node that was already in the tree will not be pruned if its depth is exactly equal to the one in the novelty table for one of its atoms (this condition appears on the third line of \textit{Check\_novelty}).

\algrenewcommand\algorithmicindent{0.5cm}

\begin{algorithm}[!t]
\caption{Planning step of $\pi$-IW(1)}
\label{algo}
\begin{algorithmic}
	\Function{Lookahead}{tree}
    \State \textit{Initialize\_labels}(tree)
	\State D := \textit{Make\_empty\_novelty\_table}()
 	\While{within\_budget \textbf{and} $\neg$tree.root.solved}
    	\State n, a := Select(tree.root, D)
        \If{a $\neq \; \perp$}
			\State Rollout(n, a, D)
        \EndIf
    \EndWhile
    \EndFunction
    
    
    \Function{Select}{n, D}
    \Loop
    	\State novel := Check\_novelty(D, n.atoms, n.depth, \textbf{false})
    	\If{is\_terminal(n) \textbf{or} $\neg$novel}
        	\State \textit{Solve\_and\_propagate\_label}(n)
        	\State \textbf{return} n, $\perp$
    	\EndIf
    	\State a := \textit{Sample\_action}(n)
    	\If{n[a] \textbf{in} tree}
    		\State n := n[a]
    	\Else
    		\State \textbf{return} n, a
    	\EndIf
    \EndLoop
    \EndFunction
    
    
    \Function{Rollout}{n, a, D}
    \While{within\_budget}
		\State n := expand\_node(n, a)
		\State n.solved := \textbf{false}

		\State novel := Check\_novelty(D, n.atoms, n.depth, \textbf{true})
    	\If{is\_terminal(n) \textbf{or} $\neg$novel}
    		\State \textit{Solve\_and\_propagate\_label}(n)
        	\State \textbf{return}
    	\EndIf
        \State a := \textit{Sample\_action}(n)
    \EndWhile
    \EndFunction
    
    
    \Function{Check\_novelty}{D, atoms, d, is\_new}
	\State novel := \textbf{false}
	\For{f \textbf{in} atoms}
    	\State novel := novel $\lor$ d $<$ D[f] $\lor$ ($\neg$is\_new $\land$ d = D[f])
    	\If{d $<$ D[f] $\land$ is\_new}
            \State D[f] := d
        \EndIf
    \EndFor
    \State \textbf{return} novel
    \EndFunction
\end{algorithmic}
\end{algorithm}

The only difference between Rollout IW and $\pi$-IW is how the function \textit{Sample\_action} is defined. In the original Rollout IW, \textit{Sample\_action} returns an action sampled with uniform probability, whereas $\pi$-IW uses a softmax policy $\widehat\pi_\theta(a|s_n) \propto \exp\left(h_a(s_n,\theta)/\tau\right)$, where $h_a$, $a\in A$, are the output logits of the NN and $\tau$ is a temperature parameter. Our planning step thus becomes the original Rollout IW in the limit $\tau\rightarrow \infty$. Just as in Rollout IW, actions that lead to nodes labelled as solved should not be considered. Thus, we set probability $\widehat\pi_\theta(a|s_n) = 0$ for each solved action $a$ and normalize $\widehat\pi_\theta$ over the remaining actions before sampling.

Every time a node is labelled as solved, we try to propagate the label along the branch to the root (\textit{Solve\_and\_propagate\_label}). Each node of the branch will be marked as solved if all of its children appear as solved. Thus, the propagation of the label stops when at least one child has not yet been pruned. Initially, all nodes of the cached tree are marked as not solved, except for the ones that are terminal (\textit{Initialize\_labels}).

\subsection{Learning Step}

Once the tree has been generated, the discounted rewards are backpropagated to the root: $R_i = r_i + \gamma \max_{j \in children(i)}{R_j}$. In general, a target policy $\pi^{target}_t(\cdot|s_t)$ can be induced from the returns at the root node by applying another softmax function, although in our experiments we applied the deterministic version (with $\tau \rightarrow 0$). The state $s_t$ is stored together with the target policy in a dataset to train the model in a supervised manner.
We use the cross-entropy error between the induced target policy $\pi_{t}^{target}(\cdot|s_t)$ and the current policy estimate $\widehat\pi_\theta(\cdot|s_t)$ to update the policy parameters $\theta$, defining a loss function
\begin{align*}
\mathcal{L}=-\pi_{t}^{target}(\cdot|s_t)^\top\log\widehat\pi_\theta(\cdot|s_t).
\end{align*}
In our experiments, we also add an $\ell$-2 regularization term to avoid overfitting and help convergence.
The model is trained by randomly sampling transitions from the dataset. The planning and learning steps can be executed in parallel, as in AlphaZero, or sequentially.
In our experiments we choose the latter, sampling a batch of transitions at each iteration.
We keep a maximum of $T$ transitions, discarding outdated transitions in a FIFO manner.

Finally, a new root is selected from the nodes at depth 1 by selecting an action $a_t \sim \pi^{target}_t(\cdot|s_t)$, and the resulting subtree is kept for the next planning step. Cached nodes are not added to the novelty table, since it has been argued in previous work that this increases exploration and hence performance \cite{Lipovetzky2015ClassicalGames}. Note that cached nodes will contain outdated information. We did not find this to have a great impact on performance, and one possibility could be to rerun the model on all nodes of the tree at regular intervals (this is not done in our experiments).

\subsection{Dynamic Features}

The quality of the transitions recorded by IW greatly depends on the feature set $\Phi$ used to define the novelty of states. For example, even though IW has been applied directly to visual (pixel) features~\cite{bandres2018planning}, it tends to work best when the features are {\em symbolic}, e.g., when the RAM state is used as a feature vector~\cite{Lipovetzky2015ClassicalGames}.
Symbolic features make planning more effective, since the width of a problem is effectively reduced by the information encoded in the features.
However, how to automatically learn powerful features for this type of structured exploration is an open challenge.

Unlike previous width-based methods, $\pi$-IW can use the representation learned by the policy NN to define a feature space, as in representation learning~\cite{Goodfellow-et-al-2016}.
With this dependence, the behavior of IW effectively changes when interleaving policy updates with runs of IW. If appropriately defined, these features should help to distinguish between important parts of the state space. In this work, we extract $\Phi$ from the last hidden layer of the NN. In particular, we use the output of the rectified linear units that we subsequently discretize in the simplest way, resulting in binary features ($0$ for zero outputs and $1$ for positive outputs). 

\section{Experiments}\label{sec:exp}

In this section, we evaluate the performance of Policy-Guided Iterated-Width ($\pi$-IW) in different settings. First, we consider a simple problem where we compare our method against AlphaZero and current width-based methods. Second, we present results in the Atari 2600 testbed. 
The following questions are addressed: 
\begin{itemize}
\item How do the different types of exploration 
(structured for $\pi$-IW and unstructured for MCTS) affect the performance of both algorithms? 
\item What is the benefit of learning a policy to guide the IW planner? 
\item Are the learned (dynamic) features effective? Is it possible to learn them without degrading the performance?
\end{itemize}

\begin{figure}
\centering
\includegraphics[width=0.95\columnwidth]{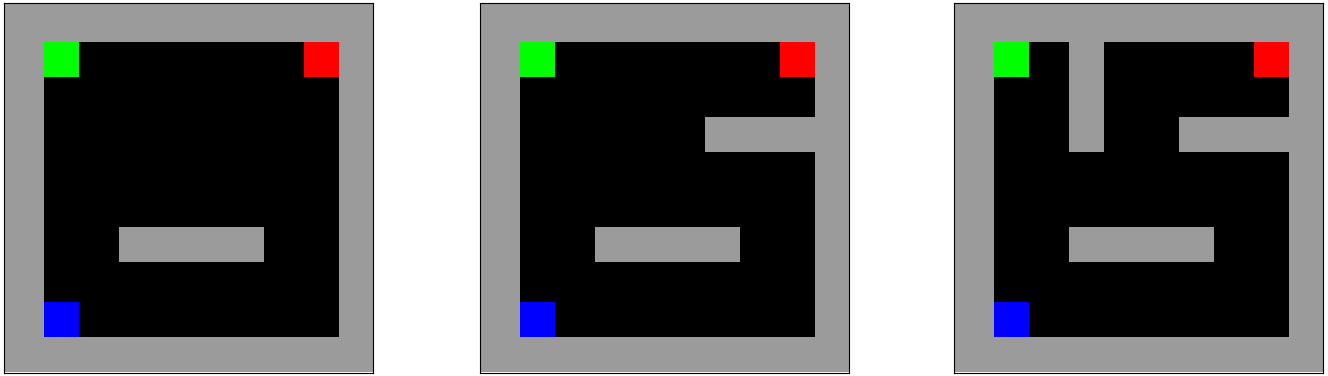}
\caption{Snapshot of three versions of the maze. The blue, red and green squares represent the agent, the key and the door, respectively.}
\label{fig:gridworld}
\end{figure}

To answer the previous questions, we define a grid-like environment where an agent (blue pixel) has to navigate to first pick up a key (red pixel) and then go through a door (green pixel). An episode terminates with a reward of $+1$ when the goal is accomplished, with a reward of $-1$ when a wall is hit, or with no reward after a maximum of 200 steps is reached. Intermediate states are \emph{not} rewarded (including picking up the key), which makes the the problem more challenging. The observation is an $84 \times 84$ RGB image and possible actions are no-op, going up, down, left or right. See Figure \ref{fig:gridworld} for an example.

We first analyze $\pi$-IW using static and dynamic features. For the first case, we take the set of BASIC features \cite{Bellemare2012TheAgents}, where the input image is divided in tiles and an atom, represented by a tuple $(i,j,k)$, is true if color $k$ appears in the tile $(i,j)$. In our simple environment, we make the tiles coincide with the grid, and we call this variant $\pi$-IW(1)-BASIC.
For the second case, we take the (discretized) outputs of the last hidden layer of the policy network as binary feature vectors. We call this variant $\pi$-IW(1)-dynamic.

\subsection{$\pi$-IW Can Reduce the Width of a Problem}

\begin{figure}
\centering
\includegraphics[width=0.8\columnwidth]{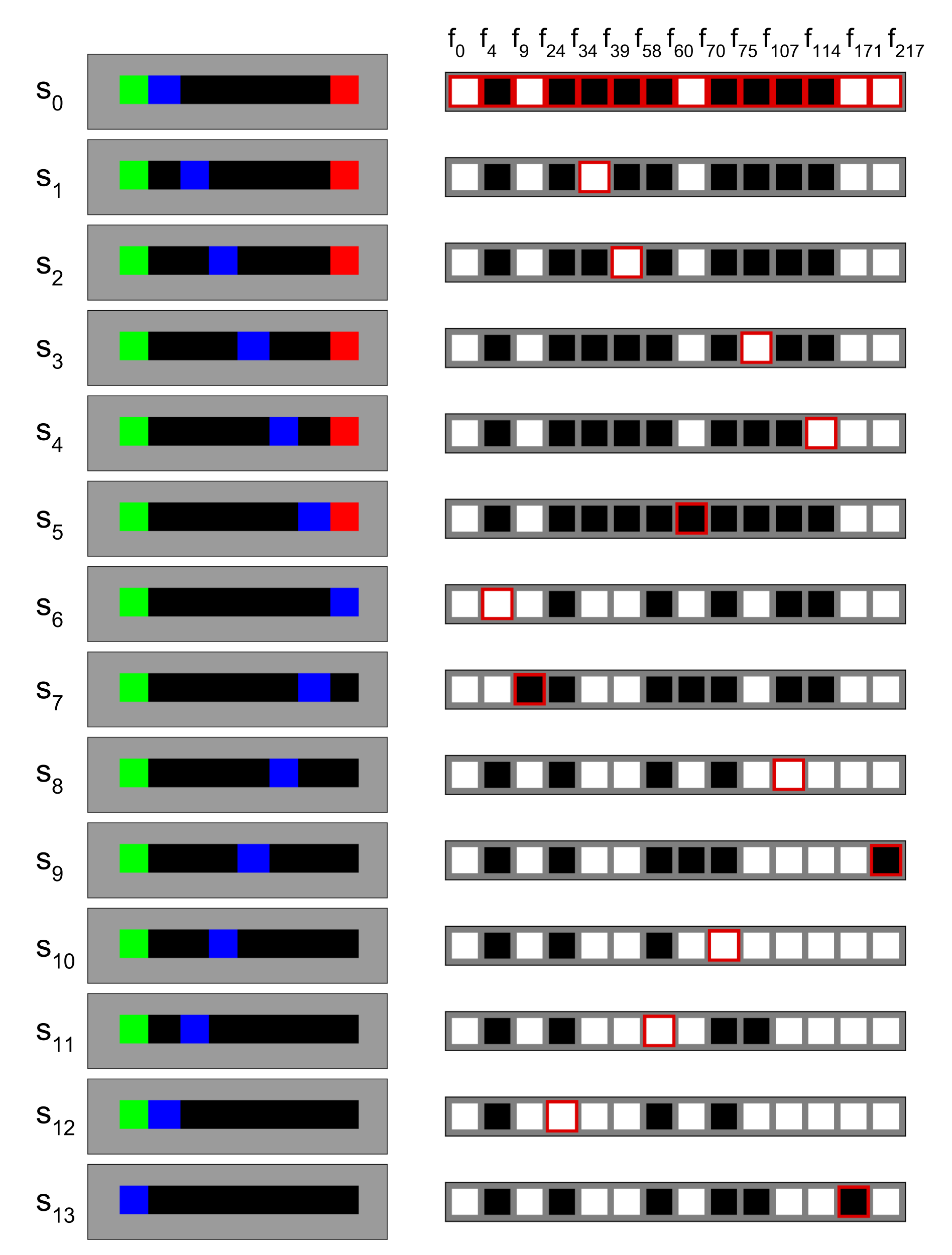}
\caption{Feature learning in a corridor task. Left: from top to bottom, states expanded by $\pi$-IW(1) in one planning step once the policy has been trained. Right: subset of learned features for each state. Novel features at each step are marked in red.}
\label{fig:corridor}
\end{figure}

Our first example is a simple corridor where the agent is located between a key and a door
(see Figure \ref{fig:corridor}). Using the BASIC features, this problem has width 2, since the agent needs to keep track of the paired features \emph{having the key} (final pixel in blue or red) and \emph{visited position} (color blue or other) jointly. Therefore, it is not solvable by IW(1) in the classical (off-line) setting.
However, in the online planning setting (where novelty tables are reset after each action execution step), the task is solvable by IW(1).
We are interested in analyzing the behavior of $\pi$-IW using dynamic features.

As expected, one planning step of $\pi$-IW using the set of BASIC features does not reach the goal, and results in a trajectory pruned at $s_8$, two steps after picking up the key. Similarly, $\pi$-IW(1)-dynamic is unable to generate the optimal plan in its first planning step, since the initial features are uninformed.
However, using a sufficient number of features (13 boolean features in this example), after learning, $\pi$-IW(1)-dynamic \emph{solves the task using a single planning (offline) step} in five out of five cases. This shows that the problem width is \emph{reduced} from 2 to 1 in the learned representation of the policy. This is remarkable, since there is no explicit term in the loss function that encourages the policy to generate such a representation. 



\subsection{$\pi$-IW Improves MCTS Exploration}

We now compare the two $\pi$-IW variants with current width-based methods and AlphaZero \cite{Silver2017MasteringChess} in a more complex task. We consider three variants of a maze, with increasing difficulty, shown in Figure \ref{fig:gridworld}.
Although AlphaZero was originally designed for two-player zero-sum games, it can be easily extended to the MDP setting.
Each time a node $s$ is generated, the statistics $W_n$ of all nodes in the trajectory from the root to $s$ are updated with the value of $s$. Since in the MDP setting there are rewards in the intermediate edges, we update $W_n$ with the discounted sum of rewards of all edges between nodes n and $s$, including the value of $s$ (e.g. $W_1 \gets W_1 + r_1 + \gamma r_2 + \gamma^2 v_3$).

AlphaZero controls the balance between exploration and exploitation by a parameter $p_{uct}$ together with a temperature parameter in the target policy $\tau$, similar to ours. In the original paper, $\tau$ is set to 1 for a few steps at the beginning of every episode, and then it is changed to an infinitesimal temperature $\tau=\epsilon$ for the rest of the game \cite{Silver2017MasteringKnowledge}. Nevertheless, we achieved better results in our experiments with AlphaZero using $\tau = 1$ for the entire episode.

\begin{figure*}[t]
\centering
\includegraphics[height=0.38\textheight]{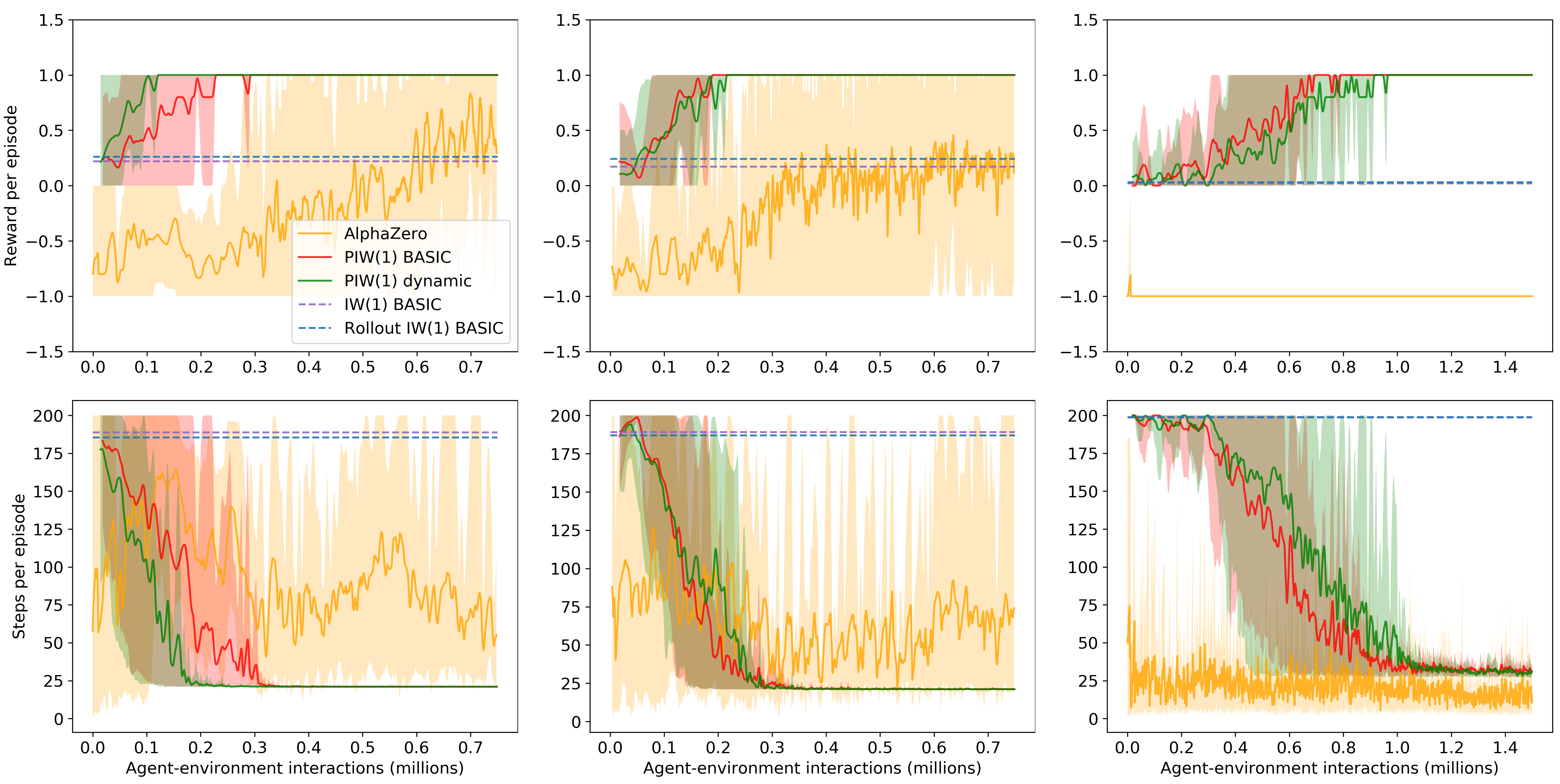}



\caption{Top: Performance of width-based planning algorithms and AlphaZero in the three simple mazes is shown in the first row, increasing difficulty from left to right (1, 2 and 3 walls respectively). Bottom: The number of transitions per episode. We plot averages over five runs for learning algorithms, and shades show the minimum and maximum values. IW and RolloutIW baseline results are averages over 100 runs.}
\label{fig:simple-performance}
\end{figure*}


\begin{table}
\begin{center}
\begin{small}
\begin{tabular}{lcc}
\toprule
Hyperparameter &  Value & Algorithm \\
\midrule
Discount factor & 0.99 & Both \\
Batch size & 32 & Both \\
Learning rate &  0.0005 & Both \\
Clip gradient norm & 40 & Both \\
RMSProp decay & 0.99 & Both \\
RMSProp epsilon & 0.1 & Both \\
Tree budget nodes & 50 & Both \\
Dataset size $T$ & $10^3$ & Both \\
L2 reg. loss factor & $10^{-3}$ & Both \\
Tree policy temp. $\tau$ & 1 & Both \\
$p_{uct}$ & 0.5 & AlphaZero \\
Diritchlet noise $\alpha$ & 0.03 & AlphaZero \\
Noise factor & 0.25 & AlphaZero \\
Value loss factor & 1 & AlphaZero \\
\bottomrule
\end{tabular}
\end{small}
\end{center}
\caption{Hyperparameters used for $\pi$-IW and AlphaZero.}
\label{tab:hyperparameters}
\end{table}

\begin{figure}
\centering
\includegraphics[width=0.95\columnwidth]{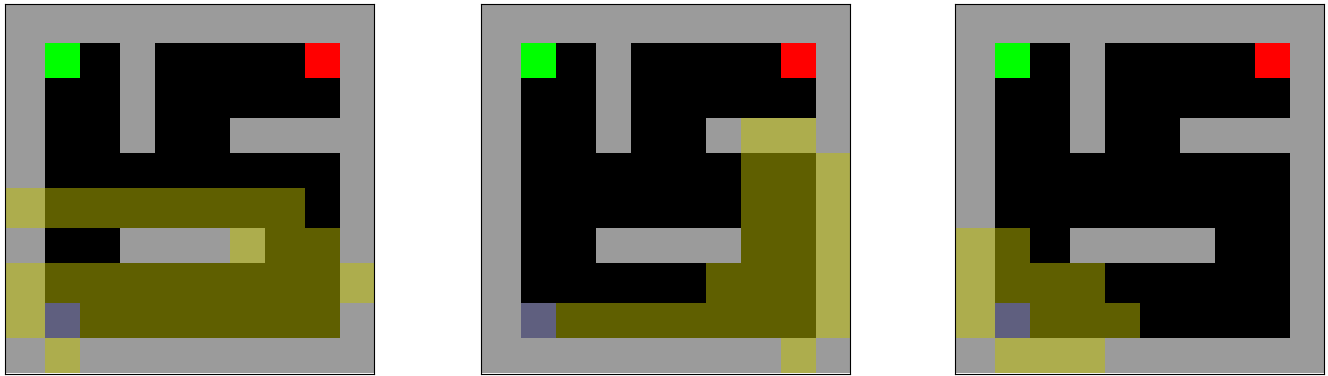}
\caption{Three illustrative examples of visited states (in yellow) during one planning step before learning: \textbf{(left)} $\pi$-IW(1)-BASIC, \textbf{(middle)} $\pi$-IW(1)-dynamic, and \textbf{(right)} AlphaZero. The $\pi$-IW variants explore a larger region of the state space than AlphaZero.}
\label{fig:heatmap}
\end{figure}

Both $\pi$-IW and AlphaZero algorithms share the same NN architecture and hyperparameters, specified in Table \ref{tab:hyperparameters}. We use two convolutional and two fully connected layers as in \citet{mnih2013playing}, which are trained using the non-centered version of the RMSProp algorithm. All hyperparameters of AlphaZero and $\pi$-IW have been optimized for the second version of the game, with two walls.

Figure \ref{fig:simple-performance} shows results comparing $\pi$-IW against existing width-based algorithms and AlphaZero for the three mazes (we also performed experiments using different random configurations of the walls and the results remained consistent). The top row shows the average reward as a function of the number of interactions with the environment.
As expected, the number of interactions required to solve the problem increases with the level of difficulty.
While $\pi$-IW variants reach top performance in less than $2\cdot 10^5$ interactions in the first and the second versions of the game, they require nearly $10^6$ to fully solve the third version.

The performance of the two other width-based algorithms (IW and Rollout IW) is independent of the number of interactions.
These algorithms, despite using the structured exploration of IW, are limited to width 1 and do not learn from previous visited states.
Consequently, only in a very few cases (depending on the tie breaking, basically), they find the correct sequence of actions.
Finally, AlphaZero shows less stable behavior and is unable to solve the most difficult scenario (rightmost plot). Comparing the two $\pi$-IW variants we observe few differences.
This is remarkable, since it indicates that in this simple maze the features used by IW can be learned easily from scratch.

Figure \ref{fig:simple-performance} (bottom row) compares the number of steps per episode as a function of the number of interactions.
For all versions of the game, we observe a decreasing trend in the $\pi$-IW variants until the optimal policy is learned.
In this case, episodes require progressively fewer steps because the learned policy converges to the optimal one.
The convergence occurs in alignment with the reward per episode (top row) and again, we observe no significant difference between $\pi$-IW(1)-BASIC and $\pi$-IW(1)-dynamic.
In contrast, AlphaZero shows a more irregular behavior. In the hardest instance, it ends up hitting a wall most of the time.

To illustrate the benefits of the structured exploration of $\pi$-IW compared to MCTS, we analyze the sample trajectories from the different algorithms after the first planning step, before any learning.
We calculate the length of the trajectories (removing repeated states).
On average,\linebreak AlphaZero produces a tree whose longest trajectory is $3.83$ (stdev$~=~2.15$). On the other hand, the longest trajectories of $\pi$-IW are $7.3$ (stdev $=2.5$) and $7.02$ (stdev $=2.25$), respectively for BASIC and dynamic features (averages over 100 runs).
Figure \ref{fig:heatmap} illustrates this. Although there is no guidance towards rewarding states for any of the three algorithms (since the NNs have not been trained yet), $\pi$-IW reaches deeper parts of the state space than AlphaZero thanks to the structured exploration. As before, we do not observe significant differences between using handcrafted features or the ones extracted from the NN. 

From these results we can draw the following conclusions.
First, existing width-based algorithms can be significantly improved by incorporating a guiding policy, as in $\pi$-IW.
Second, we have shown that for this simple problem, a small set of features can be effectively learned without degrading the performance of $\pi$-IW. 
Finally, $\pi$-IW outperforms AlphaZero because it uses the structure of the state to explore more systematically and reach deeper states. In contrast, the exploration of MCTS needs to go through an optimal branch several times to increase its probability for action selection, since the policy estimate is based on counts.

\subsection{$\pi$-IW on Atari Games}
We end this experimental section presenting results of $\pi$-IW(1)-dynamic on the pixel setting of the Atari suite. The aim of this section is to compare the performance of $\pi$-IW on a more challenging benchmark with existing width-based algorithms that use (predefined) pixel-based features. In particular, we consider IW and Rollout IW from \citet{bandres2018planning}. We do not provide results of AlphaZero in this benchmark, since our preliminary analysis showed poor performance, and the amount of hyperparameters to tune is considerably higher compared to $\pi$-IW.

We focus on a similar setting as in~\citet{bandres2018planning}, where a short budget is given to the planner, although we do not aim to plan in real time. In our case, we set the budget of expanded nodes to 100, resulting in approximately one second per transition, considering \emph{both planning and learning steps}. Note that this budget is very small compared to existing RAM-based methods, that allow 30,000 expanded nodes at each step \cite{Lipovetzky2015ClassicalGames,shleyfman2016blind}.

Table \ref{tab:atari-lookahead} shows results comparing $\pi$-IW using dynamic features with IW and Rollout IW using the B-PROST feature set. All hyperparameters are kept the same as in Table~\ref{tab:hyperparameters} except for tree budget $= 100$, $T = 10^4$, $\tau = 0.5$, and frameskip $= 15$. The inputs of the NN are the last 4 grayscale frames stacked to form a 4-channel image. Results of $\pi$-IW are an average of the last 10 episodes for 5 runs with different random seeds. Performance is measured after 20M generated nodes, i.e., interactions with the simulator (excluding skipped frames).

First, if we compare $\pi$-IW against the methods that use a budget of 0.5 seconds (1st and 3rd columns vs 5th column), we observe that $\pi$-IW systematically outperforms both IW and rollout variants (see values in blue). Only in 13 games is either IW or Rollout IW better than our method, and only in two cases are both better.
This shows that better performance can indeed be achieved at the cost of training the policy and learning the features.

Second, we observe (in bold) that $\pi$-IW outperforms all other methods in 24 games, and performance is comparable to the non-guided approaches in most other games (3rd and 5th columns vs 6th column). Remarkably, this is achieved with significantly less computational budget (approximately 30 times less) and without the need of predefined features.

These results suggest that a guiding policy can be beneficial, not only in terms of computational budget, but also in terms of the learned representation (in our case, the simple discretized features of the hidden layer) that can be directly exploited by the IW planner.

\section{Conclusions and Future Work}\label{sec:fin}

We have presented $\pi$-IW, an algorithm that effectively combines planning and learning. Our approach learns a compact policy using the exploration power of IW(1), which helps reaching distant high-reward states. We use the transitions recorded by IW(1) to train a policy in the form of a neural network. Simultaneously, the search is informed by the current policy estimate, reinforcing promising paths.

We have shown that the learned representation by the policy network can be used as a feature space for IW, removing the need of pre-defined features. Interestingly, $\pi$-IW can even learn representations that reduce the width of a task. In our simple example, the minimum number of binary features required to decrease the width was determined by the distance to a rewarded state. For complex problems with a larger horizon, one could use more features, or even consider different feature discretizations other than binary ones. We believe this is a promising line of future research.

Our algorithm operates in a similar manner to AlphaZero, except that the exploration relies on the pruning mechanism of IW, it does not keep a value estimate, and the target policy is based on observed rewards rather than visitation counts.
Compared to AlphaZero and previous width-based methods, $\pi$-IW has superior performance in simple environments. In the Atari 2600 games, $\pi$-IW achieves a similar performance to Rollout IW with a much smaller planning budget and without the need to provide pre-defined features.

In the future, we would like to investigate the use of a value estimate in our algorithm, and the possibility to decouple learning features for IW from the policy estimate. We also plan to bring $\pi$-IW closer to the RL setting by allowing stochastic state transitions.
This requires analyzing how the final policy is determined by the interplay between the planning and the learning steps.
Recent RL methods have proposed to exploit other notions of novelty for systematic exploration~\cite{machado,pmlr-v70-ostrovski17a,Martin2017Count-BasedLearning,Bellemare2016UnifyingMotivation}.
We believe that the combination of structured exploration and representation learning of $\pi$-IW is a promising direction for efficient exploration in RL.

\begin{table*}
\input{new_atari_table}
\label{tab:atari-lookahead}
\end{table*}

\section{Acknowledgements}
Miquel Junyent's research is partially funded by project 2016DI004 of the Catalan Industrial Doctorates Plan. Anders Jonsson is partially supported by the grants TIN2015-67959 and PCIN-2017-082 of the Spanish Ministry of Science.
Vicen\c{c} G\'omez is supported by the Ramon y Cajal program RYC-2015-18878 (AEI/MINEICO/FSE,UE) and the 
Mar\'ia de Maeztu Units of Excellence Programme (MDM-2015-0502).

\bibliographystyle{aaai}
\bibliography{bibliography}

\end{document}

%% file: new_atari_table.tex
\begin{center}
\begin{small}
\begin{tabular}{lrrrrrr}

\toprule
Algorithm & IW & IW & Rollout IW & Rollout IW & $\pi$-IW \\
Features & BPROST & BPROST & BPROST & BPROST & Dynamic \\
Planning horizon & 0.5s & 32s & 0.5s & 32s & \#100 \\

\midrule

Alien & 1316.0 & \textbf{14010.0} & 4238.0 & 6896.0 & 3969.8 \\
Amidar & 48.0 & 1043.2 & 659.8 & \textbf{1698.6} & \textcolor{blue}{950.4} \\
Assault & 268.8 & 336.0 & 285.6 & 319.2 & \textcolor{blue}{\textbf{1574.9}} \\
Asterix & 1350.0 & 262500.0 & 45780.0 & 66100.0 & \textcolor{blue}{\textbf{346409.1}} \\
Asteroids & 840.0 & \textbf{7630.0} & 4344.0 & 7258.0 & 1368.5 \\
Atlantis & 33160.0 & 82060.0 & 64200.0 & \textbf{151120.0} & \textcolor{blue}{106212.6} \\
Bank Heist & 24.0 & 739.0 & 272.0 & \textbf{865.0} & \textcolor{blue}{567.2} \\
Battle zone & 6800.0 & 14800.0 & 39600.0 & \textbf{414000.0} & \textcolor{blue}{69659.4} \\
Beam rider & 715.2 & 1530.4 & 2188.0 & 2464.8 & \textcolor{blue}{\textbf{3313.1}} \\
Berzerk & 280.0 & 1318.0 & 644.0 & 862.0 & \textcolor{blue}{\textbf{1548.2}} \\
Bowling & 30.6 & \textbf{49.2} & 47.6 & 45.8 & 26.3 \\
Boxing & 99.4 & 79.0 & 75.4 & 79.4 & \textcolor{blue}{\textbf{99.9}} \\
Breakout & 1.6 & 56.0 & 82.4 & 36.0 & \textcolor{blue}{\textbf{92.1}} \\
Centipede & 88890.0 & \textbf{143275.4} & 36980.2 & 65162.6 & \textcolor{blue}{126488.4} \\
Chopper command & 1760.0 & 1800.0 & 2920.0 & 5800.0 & \textcolor{blue}{\textbf{11187.4}} \\
Crazy climber & 16780.0 & 44340.0 & 39220.0 & 43960.0 & \textcolor{blue}{\textbf{161192.0}} \\
Demon attack & 106.0 & 23619.0 & 2780.0 & 9996.0 & \textcolor{blue}{\textbf{26881.1}} \\
Double dunk & -22.0 & -22.4 & 3.6 & \textbf{20.0} & \textcolor{blue}{4.7} \\
Enduro & 2.6 & 229.2 & 169.4 & 359.4 & \textcolor{blue}{\textbf{506.6}} \\
Fishing derby & -83.8 & -39.0 & -68.0 & -16.2 & \textcolor{blue}{\textbf{8.9}} \\
Freeway & 0.6 & \textbf{25.0} & 2.8 & 12.6 & 0.3 \\
Frostbite & 106.0 & 182.0 & 220.0 & \textbf{5484.0} & \textcolor{blue}{270.0} \\
Gopher & 1036.0 & \textbf{18472.0} & 7216.0 & 13176.0 & \textcolor{blue}{18025.9} \\
Gravitar & 380.0 & 1630.0 & 1630.0 & \textbf{3700.0} & \textcolor{blue}{1876.8} \\
HERO & 2034.0 & 7432.0 & 13709.0 & 28260.0 & \textcolor{blue}{\textbf{36443.7}} \\
Ice hockey & -13.6 & -7.0 & -6.0 & \textbf{6.6} & -11.7 \\
James bond 007 & 40.0 & 180.0 & 450.0 & \textbf{22250.0} & 43.2 \\
Kangaroo & 160.0 & 3820.0 & 1080.0 & \textbf{5780.0} & \textcolor{blue}{1847.5} \\
Krull & 3206.8 & 5611.8 & 1892.8 & 1151.2 & \textcolor{blue}{\textbf{8343.3}} \\
Kung-fu master & 440.0 & 8980.0 & 2080.0 & 14920.0 & \textcolor{blue}{\textbf{41609.0}} \\
Montezuma's revenge & \textbf{0.0} & \textbf{0.0} & \textbf{0.0} & \textbf{0.0} & \textcolor{blue}{0.0} \\
Ms. Pac-man & 2578.0 & \textbf{20622.8} & 9178.4 & 19667.0 & \textcolor{blue}{14726.3} \\
Name this game & 7070.0 & \textbf{13478.0} & 6226.0 & 5980.0 & \textcolor{blue}{12734.8} \\
Phoenix & 1266.0 & 5550.0 & 5750.0 & \textbf{7636.0} & \textcolor{blue}{5905.1} \\
Pitfall! & \textbf{-8.6} & -92.2 & -81.4 & -130.8 & -214.8 \\
Pong & -20.8 & 0.8 & -7.4 & \textbf{17.6} & -20.4 \\
Private eye & 2690.8 & -526.4 & -322.0 & \textbf{3157.2} & 452.4 \\
Q*bert & 515.0 & 16505.0 & 3375.0 & 8390.0 & \textcolor{blue}{\textbf{32529.6}} \\
Road Runner & 200.0 & 0.0 & 2360.0 & 37080.0 & \textcolor{blue}{\textbf{38764.8}} \\
Robotank & 3.2 & 32.8 & 31.0 & \textbf{52.6} & 15.7 \\
Seaquest & 168.0 & 356.0 & 980.0 & \textbf{10932.0} & \textcolor{blue}{5916.1} \\
Skiing & -16511.0 & -15962.0 & \textbf{-15738.8} & -16477.0 & -19188.3 \\
Solaris & 1356.0 & 2300.0 & 700.0 & 1040.0 & \textcolor{blue}{\textbf{3048.8}} \\
Space invaders & 280.0 & 1963.0 & 2628.0 & 1980.0 & \textcolor{blue}{\textbf{2694.1}} \\
Stargunner & 840.0 & 1340.0 & 13360.0 & \textbf{15640.0} & 1381.2 \\
Tennis & -23.4 & -22.2 & -18.6 & \textbf{-2.2} & -23.7 \\
Time pilot & 2360.0 & 5740.0 & 7640.0 & 8140.0 & \textcolor{blue}{\textbf{16099.9}} \\
Tutankham & 71.2 & 172.4 & 128.4 & 184.0 & \textcolor{blue}{\textbf{216.7}} \\
Up'n down & 928.0 & 62378.0 & 36236.0 & 44306.0 & \textcolor{blue}{\textbf{107757.5}} \\
Venture & 0.0 & \textbf{240.0} & 0.0 & 80.0 & \textcolor{blue}{0.0} \\
Video pinball & 28706.4 & 441094.2 & 203765.4 & 382294.8 & \textcolor{blue}{\textbf{514012.5}} \\
Wizard of wor & 5660.0 & \textbf{115980.0} & 37220.0 & 73820.0 & \textcolor{blue}{76533.2} \\
Yars' revenge & 6352.6 & 10808.2 & 5225.4 & 9866.4 & \textcolor{blue}{\textbf{102183.7}} \\
Zaxxon & 0.0 & 15080.0 & 9280.0 & 22880.0 & \textcolor{blue}{\textbf{22905.7}} \\

\midrule

\# best & 1 & 10 & 1 & 17 & 24 \\

\bottomrule

\end{tabular}

\end{small}
\end{center}
\caption{Score comparison of width-based methods in 54 Atari games (pixel setting). Scores in bold are best overall and values in blue are $\pi$-IW scores higher than methods with 0.5s of budget. Performance of $\pi$-IW is an average of the last 10 episodes for 5 runs after 20M interactions (including all generated trees). Other results are taken from Bandres, Bonet and Geffner (2018).} 